\documentclass{article}

\usepackage[preprint,nonatbib,final]{neurips_2019}
\usepackage[utf8]{inputenc} 
\usepackage[T1]{fontenc}    
\usepackage{hyperref}       
\usepackage{url}            
\usepackage{booktabs}       
\usepackage{amsfonts}       
\usepackage{nicefrac}       
\usepackage{microtype}      
\usepackage{graphicx}
\usepackage{float}

\title{Sparse associative memory based on contextual code learning for disambiguating word senses}

\author{
  Max Raphael Sobroza \\
  Institut Mines-Télécom Atlantique\\
  \texttt{max.sobrozamarques@imt-atlantique.fr} \\
  \And
  Tales Marra \\
  Institut Mines-Télécom Atlantique\\
   \texttt{tales.marra@imt-atlantique.net} \\
    \And
    Deok-Hee Kim-Dufor \\
  Centre Hospitalier Universitaire de Brest\\
  \texttt{dh.kimdufor@gmail.com} \\
  \And
  Claude Berrou \\
   Institut Mines-Télécom Atlantique\\
   \texttt{claude.berrou@imt-atlantique.fr}\\ 
}

\begin{document}

\maketitle
\begin{abstract}
  In recent literature, contextual pretrained Language Models (LMs)  demonstrated their potential in generalizing the knowledge to several Natural Language Processing (NLP) tasks including supervised Word Sense Disambiguation (WSD), a challenging problem in the field of Natural Language Understanding (NLU). 
  However, word representations from these models are still very dense, costly in terms of memory footprint, as well as minimally interpretable.
 In order to address such issues, we propose a new supervised biologically inspired technique for transferring large pre-trained language model representations into a compressed representation, for the case of WSD. 
 Our produced representation contributes to increase the general interpretability of the framework and to decrease memory footprint, while enhancing performance. 
  
\end{abstract}
\section{Introduction}

Recently, large pretrained LMs such as ELMo \cite{Peters:2018} or BERT \cite{devlin-etal-2019-bert} on large corpora achieved state-of-the-art performance on several downstream tasks in NLP. These models are trained to predict the next word given a sequence of words or a masked word given it surrounding words in the same sentence. These architectures are capable to incorporate into latent distributed representations the contextual information from different timescales. From the biological point of view, there are strong evidences that hierarchical cortical structures implement a similar mechanism, where the spatiotemporal pattern completion process is done recursively in order to extract the relevant information \cite{himberger2018principles,huth2016natural}. When it comes to performing  WSD, literature indicates several supervised techniques, where different types of classifiers can learn based on annotated datasets.
A particularly interesting approach to address this task is the cognitive one, where the phenomenon of WSD is treated by modeling the cognitive process through \textit{ACT} (Adaptive Control of Thought) \cite{anderson2013architecture} combined with \textit{RACE/A} (Retrieval by Accumulating Evidence in an Architecture) \cite{van2012race}, and such a modeling allows the outperforming of many systems for the task at hand \cite{dutta2012cognitive}. 

The ideas that are presented in this paper with a particular application to contextual word representations are inspired by recent studies about physicochemical properties of the brain.  As explained  in  \cite{berrou2014information}, the brain is a very noisy medium. More precisely, according to \cite{laughlin2003communication} and \cite{mazzoni2007dynamics} respectively, neurons are subject to high erasure effects on their inputs (neurotransmitters are temporarily unavailable in a synapse to respond properly to an incoming spike) and are also the source of numerous spurious (non stimulated) spikes. Because of this fluctuating behaviour it is difficult to accept that mental pieces of information could be borne by precise levels of neuronal activity. 

A way to overcome noise and inconstancies is to make information be expressed by assemblies of neurons instead of individual neurons. Associative memories can be regarded as a good means to store pieces of information with high robustness towards erasures\cite{himberger2018principles}. A strategy to combat uncertainty about neuronal activity is to materialize information by relative values instead of absolute ones. For instance, this is the way that sparse associative memories based on neural cliques \cite{gripon2011sparse} are recovered through the Winner-Takes-All (WTA) mechanism which has shown high biological plausibility \cite{feldman1982connectionist}. Furthermore, the WTA principle is applied in several studies \cite{makhzani2015winner,pmlr-v28-goodfellow13,shu} for increase activation sparseness in neural networks.

The simplest way to implement the WTA mechanism is to organize neurons into groups or clusters, thus making it possible to achieve simple pairwise comparisons between neurons of the same cluster. This rationale was applied in \cite{srivastava2013compete,pmlr-v28-goodfellow13}. With this paper, we use an implementation based on the WTA mechanism in order to produce new contextual word representations,that are overcomplete, compositional, discrete and sparse.

\section{Related Work}

Starting from the biological perspective, by the comprehension of brain access schemes, it can be infered that the combination of binary synaptic weights, sparsely encoded memory patterns and local learning rules is one way of producing good representation \cite{palm1980associative,hacene2019budget}. And as fast information retrieval remains a critical point to the real world applications, such a method proves again its power as the computation made is basically a vector-matrix product (in the binary case this can still be considered just counting), followed by an operation of threshold \cite{palm2013neural}.

Recent work has started to address as well the problematic of memory footprint when learning word embeddings\cite{tissier2019near,shu}. In order to mitigate such an issue, \cite{shu} proposed a method that learns multi-codebook. Therefore, each word is now represented by a hash code of discrete numbers, that have to be defined in a way that similar words will have similar representations as well as a factor of difference, to capture nuances. Presenting a different use for the previous work done by the field of codebooks compression based source coding, known as product quantization (PQ) \cite{jegou2010product} and additive quantization \cite{babenko2014additive}, they show that by minimizing the squared distance between both distributions (baseline and composed embeddings), and using a direct learning approach for the codes in an end-to-end neural network, with a Gumbel-softmax layer \cite{jang2016categorical} to encourage the discreteness \cite{shu}, it is possible to construct the word embeddings radically reducing the number of parameters without hurting performance. These word codebooks are learned using a local WTA rule in the hidden layer of an autoencoder neural network.

Other works have also focused in accomplishing the task of modeling complex characteristics of word use, and its variations across different contexts. \cite{Peters:2018} presented a new type of word representation, where word vectors are learned functions of the internal states of a deep bidirectional language model (biLM), which is pretrained on a large text corpus \cite{Peters:2018}. All those aspects allow richer word representations, and also the capture of important context dependant aspects. And as the computation is made based on the internal states of the baseline architecture, such approach can be easily integrated to existing models. Results have shown ELMo's capacity to boost performance in many NLP tasks, including WSD, using the approach of the 1-nearest neighbor, similarly to \cite{melamud2016context2vec} and accordingly to the framework of evaluation from \cite{raganato2017word}.
 
 As stated above, when the vocabulary is limited, the problem of compressing word embedding representations by contextual compositional codes is already addressed. However, the problem of compressed context representation remains, as a result of the enormous variability of contexts for a single word. With the current work, our goal is to treat this problem in the particular case of WSD.
 
\section{Our Framework}
\begin{figure}[H]
\centering
\includegraphics[scale=0.16]{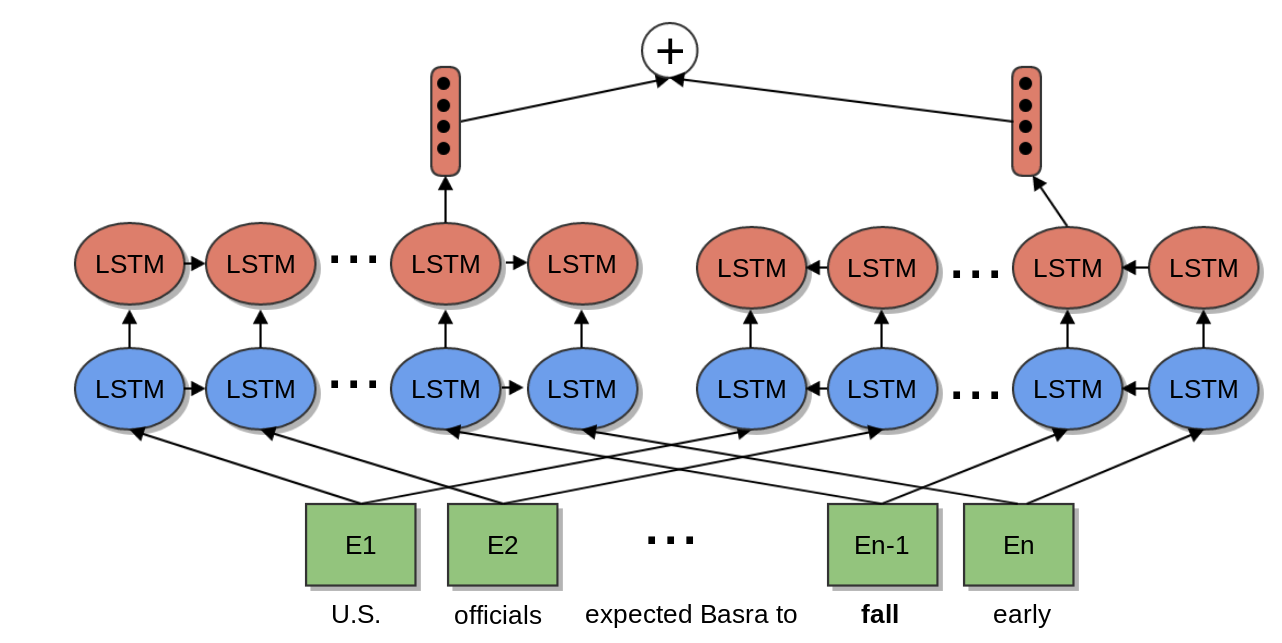}
\caption{2-layer Bidirectional LM architecture of ELMo feature extractor used for WSD task. The input sentence in this example is \textit{U.S officials expected Basra to fall early} and the ambiguous word is \textit{fall}. The output is the concatenated vector (with dimension 1024) of forward and backward latent vectors from the last layer of the feature extractor.}\label{elmo}
\end{figure}

\begin{figure}[H]
\centering
\includegraphics[scale=0.14]{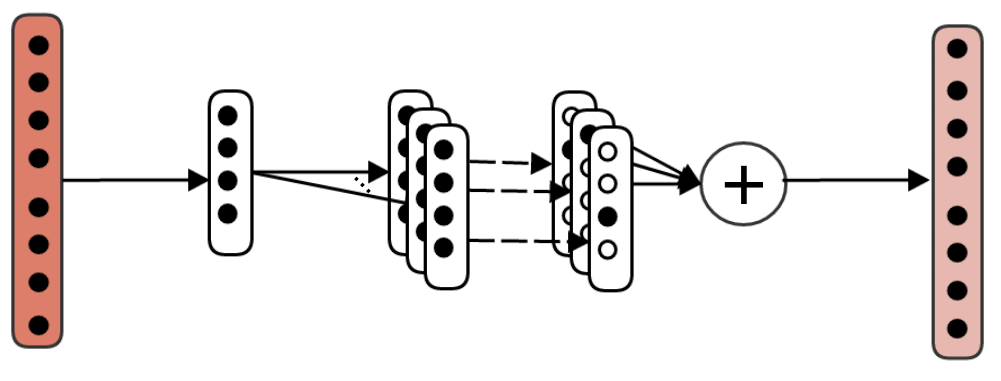}
\caption{Deep Autoencoder Neural Network Architecture for Compositional Code Learning. Solid arrows are linear projections followed by non-linear functions and dashed arrows represent the paths where \textit{Gumbel Softmax function} is applied. The input is a context vector extracted from first step and the output is the reconstructed vector. The third intermediate layers are the extracted contextual code vector (with dimension $K$ times $M$), with $K$ being the number of neurons per cluster and $M$ being the number of clusters. Differently from \cite{shu}, our method does not use the reconstruction part of the neural network.}\label{ccl}
\end{figure}

The proposed method is divided into three different steps. In the first step, the features corresponding to the ambiguous word are extracted from the second layer of the ELMo model as pictured in Fig. \ref{elmo}. Experimental results from \cite{Peters:2018} indicate that performing such extraction on the last layer of the LM (concatenating the representations from both directions) increase performance in comparison with doing it on the first one. For the case of composed words (i.e. Mother-in-law), the average embedding is taken, in order to create a more robust representation. 
In the second step, a deep autoencoder neural network model is optimized to reduce the mean squared error of reconstructed contextual features extracted from first step. This model implements the \textit{Gumbel Softmax} reparametrization function \cite{jang2016categorical} to obtain discrete activations in the last intermediate layer. For the sake of simplicity, we adopted the same architecture as \cite{shu}, as shown in Fig. \ref{ccl}.

The third step relies on \textit{Sparse Associative Memories} (SAM), neural networks able to store and retrieve sparse patterns from incomplete inputs. 
Our choice to this technique comes from the fact the sparse projections produced by SAM's have been shown to be sufficient to uniquely encode a neural pattern \cite{sam}. The connections in sparse associative memories are completely binary (they either exist or not). At the beginning of training procedure, the graph that models the SAM is initialized without any connection. Each node in the bottom-layer corresponds to a specific context usage of a word. Thus, the learning procedure consists in adding connections to the network for each pattern to store. 
The retrieving procedure starts with partial pattern, meaning that some of its active neurons are not activated in the beginning. Then, it tries to find in each group of neurons, the neuron (or the neurons) that has the maximum number of connections with the active neurons (WTA algorithm).
\begin{figure}[H]
\centering
\includegraphics[scale=0.20]{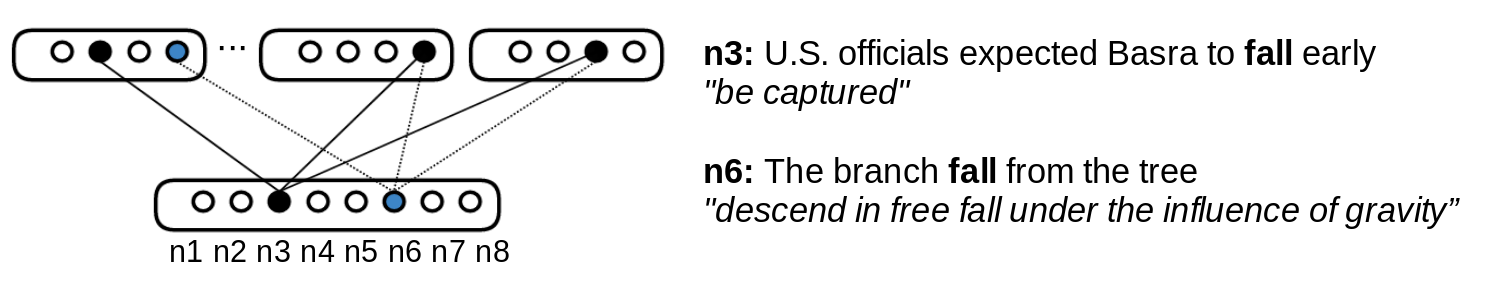}
    \caption{Example of two stored patterns in the SAM of the same word. The top layer represents input patterns extracted from compositional code features. Each node in the bottom layer is a different context application of the word \textit{fall}. For instance, the node \textit{n3} represents a context that \textit{fall} is assign to the meaning \textit{be captured} and the node \textit{n6} represents a context with the meaning \textit{descend in free fall under the influence of gravity} of the word \textit{fall}.}
\end{figure}
\section{Results}
With the purpose of learning robust discrete representations in first step, we have used all sentences and word contexts from the \textit{One billion word} dataset \cite{chelba2014one} to train the autoencoder model. In order to make our framework comparable to the baseline methods, we have maintained the process of evaluation done by \cite{raganato2017word}. Compression rate values were computed considering the saved space when storing all WSD training ELMo contextual vectors. As detailed in Table 1, the framework can achieve high compression rates, while increasing performance.

\begin{table}
\begin{center}
\begin{tabular}{ c c c c } 
Method & $K$ times $M$ & Compression Rate & F1-score \\
\hline
context2vec \cite{melamud2016context2vec} & -- & -- & 67.9 \% \\
ELMo \cite{Peters:2018} & -- & -- & 69.0 \% \\
Ours & 64 x 128 & 42.6x & \textbf{73.4 \%}  \\ 
Ours & 64 x 64 & 85.3x & 72.7 \% \\ 
Ours & 32 x 64 & 102.4x & 72.4 \% \\ 
Ours & 32 x 32 & \textbf{204.8x} & 72.5 \% \\
\hline
\label{tab1}
\end{tabular}
\caption{Table 1: Baseline and variations of our approach with their respective compression rate and F1-score in supervised WSD evaluation}
\end{center}
\end{table}

 In terms of interpretability, \cite{faruqui2015sparse} point that sparse and overcomplete word vector representations are more interpretable than dense ones.
 
 \begin{figure}[H]
\centering
\includegraphics[scale=0.6]{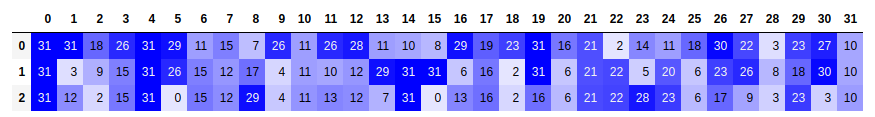}
\caption{Compositional codes ($K$=32 and $M$=32) of following contexts of the word \textbf{rock}: (0) \textit{the exhibit was inaugurated by a Colombian \textbf{rock} band , featuring a guitar...}, (1) \textit{Sand and \textbf{rock} are quarried ... in construction projects ...}, (2) \textit{Demand for \textbf{rock}, sand and gravel ... on natural resources}. Each column represents a different cluster and each cell value represents the position of the activated neuron inside a cluster.}
\label{rock}
\end{figure}
 
 In Figure \ref{rock}, the last two contexts usages have both similar meaning whereas the first context has different meaning from others, and this result is congruent to the hamming distance between different compositional codes: (0) and (1) is 26 ; (1) and (2) is 16; (0) and (2) is 25. Discrete activations of contextual codes are human readable. Therefore, it is possible to affirm that the premise of interpretability from \cite{faruqui2015sparse} remains true in the case of contextual code representations.

\section{Conclusion}
We have presented a technique of transfer learning of contextual LM representations based on neuro-inspired sparse binary associative memories.
We have confirmed by our results that such a framework can dramatically mitigate the memory footprint problem and enhance the interpretability while it is still capable of increasing performance in the focused task, WSD. Our framework is an extension of \cite{shu} work to deal with latent representations of contextual pretrained LM. For future work, we intend to explore other LM recent architectures such as BERT \cite{devlin-etal-2019-bert}, and also aggregate information from knowledge-based corpus such as WordNet \cite{miller1995wordnet}.

\section{Acknowledgements}
The authors would like to thank CAPES, NVIDIA and ERC grant for the financial support.
\small
\bibliography{sample}

\medskip

\end{document}